%% file: main.tex
\documentclass{article}

\usepackage{arxiv}
\usepackage{hyperref}
\usepackage[utf8]{inputenc} 
\usepackage[T1]{fontenc}    
\usepackage{hyperref}       
\usepackage{url}            
\usepackage{booktabs}       
\usepackage{amsfonts}       
\usepackage{nicefrac}       
\usepackage{microtype}      
\usepackage{graphicx}
\usepackage{tabularx}
\usepackage{multirow}
\usepackage{parskip}
\usepackage{float}
\usepackage{subcaption}
\usepackage{graphicx}
\usepackage{enumitem}
\usepackage{amsmath} 
\usepackage{cite}

\title{PCA-RF: An Efficient Parkinson's Disease Prediction Model based on Random Forest Classification}
\date{}   
\renewcommand{\shorttitle}{An Efficient Parkinson's Disease Prediction Model based on Random Forest Classification}

\author{ \href{https://orcid.org/0000-0003-3746-6034}{\includegraphics[scale=0.06]{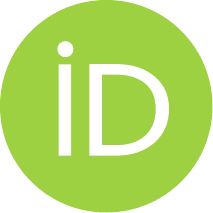}\hspace{1mm}Ishu Gupta*}
	\\
	Cloud Computing Research Center\\
	Department of Computer Science and Engineering\\ 
	National Sun Yat-sen University\\
	Kaohsiung, Taiwan\\
	\texttt{ishugupta23@gmail.com} \\
	\And
	Vartika Sharma \\
	Department of Computer Applications\\
	National Institute of Technology\\
	Kurukshetra, India \\
	136119\\
	\texttt{vsvartika.12sharma@gmail.com}
	\And
	Sizman Kaur\\
	Department of Computer Applications\\
	National Institute of Technology\\
	Kurukshetra, India \\
	136119\\
	\texttt{sizmankaur22@gmail.com}
	\And
	\href{https://orcid.org/0000-0002-8053-5050}{\includegraphics[scale=0.06]{orcid.pdf}\hspace{1mm}Ashutosh Kumar Singh} \\
	Department of Computer Applications\\
	National Institute of Technology\\
	Kurukshetra, India \\
	136119\\
	\texttt{ashutosh@nitkkr.ac.in} \\
}

\hypersetup{
pdftitle={A template for the arxiv style},
pdfsubject={q-bio.NC, q-bio.QM},
pdfauthor={David S.~Hippocampus, Elias D.~Striatum},
pdfkeywords={First keyword, Second keyword, More},
}

\begin{document}

\maketitle
\begin{abstract} 
	In this modern era of overpopulation disease prediction is a crucial step in diagnosing various diseases at an early stage. With the advancement of various machine learning algorithms, the prediction has become quite easy. However, the complex and the selection of an optimal machine learning technique for the given dataset greatly affects the accuracy of the model. A large amount of datasets exists globally but there is no effective use of it due to its unstructured format. Hence, a lot of different techniques are available to extract something useful for the real world to implement. Therefore, accuracy becomes a major metric in evaluating the model. In this paper, a disease prediction approach is proposed that implements a random forest classifier on Parkinson's disease. We compared the accuracy of this model with the Principal Component Analysis (PCA) applied Artificial Neural Network (ANN) model and captured a visible difference. The model secured a significant accuracy of up to 90\%.
\end{abstract}


\section{Introduction}
According to the fact sheets provided by the World Health Organization (WHO), the major cause that leads to disability and death all over the world is chronic diseases. Non-communicable diseases (NCD) kill around 40 million people each year, which is equal to 71\% of all deaths worldwide. Heart diseases account for most of the NCD deaths, followed by cancers, respiratory diseases, and diabetes. This led to economic output loss of 47 US trillion dollars in the previous two decades. This loss represents globally around 75\% of gross domestic product (GDP) in 2010 \cite{Kundu}. A study showed that in India around 25\% of families with a member of cardiovascular disease and 50\% of families suffering from cancer experience severe disastrous expenses and 10\% and 25\%, respectively, are affected due to poverty. Most of the estimations proved that the NCDs in India account for an overall economic loss in the range of 5-10\% of GDP, which is a significant number and is thus slowing down GDP causing a loss in development \cite{Godha}.

People in India have to spend more on treatment due to the lack of medical facilities and limited access to health insurance \cite{Treating,IJNSA}. A lot of people do not buy insurance at the early stage of their life and then repent at an older age \cite{Tiwari,Ankit}. That’s why it is very important to have an emergency fund saved separately. The healthcare issue of these types of diseases is crucial in other parts of the world as well \cite{Holistic,Sharma}. In America, over \$10,000 per person is spent out of the pocket annually on health care, more than any other country in the world \cite{Waters,DT-ILIS}. The Partnership to Fight Chronic Disease estimates that around 83 million people in the U.S. will have 3 or more chronic health problems by 2030. Some behavioral factors including unhealthy eating, lack of exercise, excessive smoking, and use of alcohol are the main reasons for these chronic diseases \cite{OnILIS,Ayushi}. Therefore, it is essential to conduct risk analysis for chronic diseases. With a great increase in the world's population, it would decline our life quality if we still depend on traditional systems of healthcare service \cite{HISA-SMFM,Kesharwani}. Such traditional systems of observing the real-time health condition of the person provided real quick assistance but with the rise in the number of data sources, there has been a flood of data in the healthcare sector \cite{JISE,Rajat}.

Due to such a big size of data records, it becomes very difficult to use the existing traditional techniques for effective results \cite{GUIM-SMD,Jalwa}. Since Big data is a recent technology in the real world that can bring large benefits to business organizations, it becomes really necessary that various types of challenges and problems associated with adopting this technology are brought into the light \cite{Harsh,EPS}. Detection of chronic diseases on early-stage helps in the early inception of preventive measures and on time-effective treatment at an initial stage has always been a better help for patients \cite{Preetesh}. Currently, maintaining clinical data sets has become a crucial step in the medical field \cite{IJAST,Kaur2018}. The patient data that holds varied features and diagnostics related to a specific disease should be inserted in the dataset with the utmost care to provide the best quality results and services \cite{Jadon}. The data entered manually in medical data sets can contain a great amount of incomplete data and redundant values, mining such healthcare data becomes time-consuming \cite{Hybrid}. As it can change the output of mining, it is important to incorporate good data preparation and data extraction prior to applying data mining algorithms. Prediction of the disease becomes faster and more efficient if data is accurate, consistent, and free from noises \cite{Kamal}. 

In this era of data explosion, a large amount of medical data is generated and updated daily \cite{Jain}. The healthcare record, as well as any electronic variant of the traditional files, contains a proper identification of the patient \cite{MACI,CC}. Medical data includes Electronic Health Records (EHR) which consists of clinical reports of patients, test reports from diagnosis, a prescription from a doctor, information from a pharmacist, information related to a person’s health insurance, social media posts such as blogs, tweets \cite{Kumar,Sloni,Animesh}. While the healthcare sector is moving towards using even more data sets into the daily routine diagnosis of patients suffering from chronic diseases, big data has truly become an effective tool in improving the patient's treatment experience. Basic machine learning techniques with the assistance of big data technology have brought forward the concept of prediction in the medical sector \cite{singh2020survey}. Machine Learning capabilities have changed healthcare in various ways, improving diagnosis of treatment choices \cite{JCOMSS,SELI,Nishad}. The predictive analysis implies doctors focus more on services and patient care. Prediction using classical models required data set of patients with disease and initial disease risk models used supervised machine learning techniques for training data \cite{BatraGarima,Arora}. These models are helpful in clinical situations and are still studied globally \cite{Kaur2017,IOSR}. The features affecting a particular disease were selected through the doctor’s experience but they could not satisfy the changes in the disease later \cite{Khushbu,IDS}. 

In this paper, we proposed a \textbf{P}rincipal \textbf{C}omponent \textbf{A}nalysis based on \textbf{R}andom \textbf{F}orest (PCA-RF) model where we applied PCA on the medical data to extract the relevant features from the data set and along with that random forest algorithm is implemented as a classifier model. Principal Component Analysis (PCA) is an unsupervised learning algorithm employed for dimensionality reduction in machine learning. It is a statistical process that converts the observations of correlated features into a set of linearly uncorrelated features with the help of orthogonal transformation.  Further, the accuracy of the model along with specificity and sensitivity is calculated. Finally, a comparison between the PCA-based Artificial Neural Network (ANN) and PCA-RF model is performed.

\section{Related Work}
The medical data contents can be maintained within the electronic format including images and may contain the patient’s identifiable details, like – films, digital imprints, or written conclusive outlines or interpretable findings. Medical data is generally represented in images or texts such as ECG, prescriptions, and MRI. Various researchers have suggested different technologies for feature selection from medical data for solving various problems like classification, regression, and retrieval. The prediction has various applications such as workload \cite{PCS,MLPAM,ICCNSJapan}, security \cite{Saxena,Confidentiality,kaur2017comparative} and much more. In \cite{Vinitha}, the projected system used delivers a deep learning technique for efficient prediction of multiple distinctive diseases occurring in vulnerable areas which hold the high frequency of diseases. \lq\lq It experimented with the modified estimate models over real-life medical data gained. It used a latent factor prototype to complete the missing data. It experimented on a regional chronic illness of cerebral infarction. It used machine learning Decision Tree algorithm and Map Reduce algorithm for data partitioning on structured and unstructured data taken from the hospital.\rq\rq Compared to various crucial estimate algorithms, the accuracy of this proposed model reached 94.8\%.

Various researchers have used different machine learning algorithms for feature extraction, classification of other chronic diseases such as Alzheimer disease \cite{Alberdi,Kruthika}, heart disease \cite{Panda,Ching-seh,Singh}, diabetes \cite{Jerjawi,Marie-Sainte}. In this paper, researchers used Indian Diabetes data taken from the UCI Repository. The system was implemented in Matlab. The Pima Indian Diabetes data set comprises approximately 768 real-life instances. The dataset contains the patient’s summary and history and the output was predicted either as positive or negative. By analyzing the performance, it was perceived that among all the algorithms that were used for training, the Levenberg-Marquardt Algorithm gave the best results based on the epochs. In paper \cite{Saji}, a model was proposed to prove the outcome of a deep learning technique and to diagnose the heart disease the data set attained from the University of California, Irvine was used. The deep learning model was analyzed on the basis of performance and further was compared with the other four efficient machine learning algorithms for predicting the status of disease from data containing a record of 566 patients from a two-record set taken from the UCI database. This learning model accomplished an accuracy score of approximately 94\% and an AUC score obtaining 0.964 in comparison to other models. The performance of this model and the non-linear machine learning algorithms was way better when compared to linear machine learning prediction models on increasing the data set size. 

In the paper, \cite{Purwar}, a hybrid learning prediction model was implemented with the help of missing value imputation (HPM-MI) that analyzes different techniques using K-means clustering and then applied with the most optimal imputation to a data set. This model was the first used amalgamation of K-means clustering along with Multi-layer Perceptron. To validate class labels for given data before applying classifier Kmeans clustering is used. This proposed system had greatly improved the quality of data by the use of the best imputation technique after the study of different eleven approaches. \lq\lq The model is evaluated on the basis of prediction and classification system and is investigated on three criteria of medical data such as Pima Indians Diabetes, Wisconsin Breast Cancer, and Hepatitis dataset from the UCI Repository.\rq\rq With respect to the performance matrix, the specificity, accuracy, sensitivity; kappa statistics, and the area under ROC were evaluated for best possible outcomes.

In paper \cite{Chaurasia}, the investigation was done on the basis of the performance of different classification models. The breast cancer dataset which had 683 instances and 10 features were used for testing, based on classification accuracy. The breast cancer data found from the Wisconsin data set from the UCI repository was analyzed with the intention of developing an efficient prediction classifier for breast cancer disease with the help of various data mining approaches. In this observation, there were three classification methods that were implemented and comparison results showed that the Sequential Minimal Optimization (SMO) had a greater accuracy i.e. 96.2\% than IBK and BF methods. There are a lot of other techniques used by researchers for predicting Parkinson’s disease \cite{Rastegar,Grover,Challa,Agarwal,Gokul,Naghavi}. In this paper, we proposed a PCA-based decision tree model for the diagnosis of chronic disease at an early stage. Then, in the end, we have compared the ANN and Random Forest techniques for the same data set.  

\section{Proposed System}
The proposed PCA-RF model uses a decision tree-based random forest algorithm in the core. In addition to that, it also exploits PCA that combines a large number of features into a comparatively small set of new principal components. Further, we pass these reduced features into our classifier that predicts the occurrence of disease. Finally, a comparative study is done using two classifying techniques – PCA along with ANN and PCA along with Random Forest based on various parameters. Fig. 1 represents the overall process of the proposed technique.

\begin{figure}[h]
   \centering
    \includegraphics[width=0.75\textwidth]{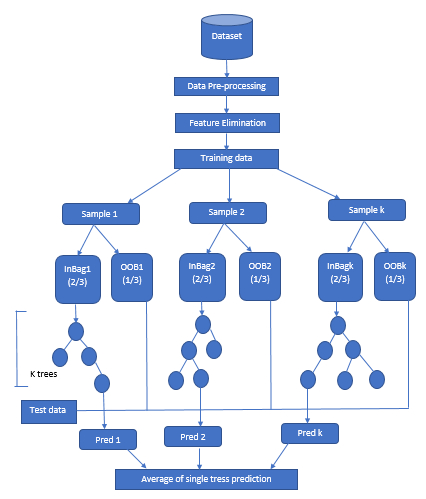}
    \caption{The workflow of the proposed disease prediction scheme}
    \label{fig:my_label6}
\end{figure}

\subsection{Feature Extraction}
More features can decrease the accuracy of the model since there is more data that needs generalization. To reduce the model’s complexity and avoid over-fitting of data, feature selection or feature extraction can be used. Feature extraction deals with deriving information from the data set containing a large number of features to construct a new feature sub-space. Here, the PCA algorithm is used for feature extraction.  PCA finds the direction of greatest variance in higher dimension data set and projects it onto a new sub-space with fewer dimensions than the original one. Mathematically, the covariance matrix for the available data is computed from the mean-subtracted data matrix in Eq. (1). Then, the Eigen vector-matrix V (must be unit Eigen-vectors) which diagonalizes C is calculated in Eq. (2), where D represents the diagonal matrix for Eigen-values of C. The Eigen-vector with the maximum Eigen-value is the principal component of the data. The corresponding Eigen-vectors give the components in order of their importance. The reduced set of principal components is passed to the classifier for prediction.

\begin{equation}
    C=\frac{1}{n-1}B*B
\end{equation}
\begin{equation}
    V^{-1}CV=D
\end{equation}

\subsection{Network Training}
Random Forest is a prototype consisting of an ample number of decision trees. This model makes use of two major theories suggested the name \lq\lq random \rq\rq. First, a random specimen of trained data points is done on constructing trees. The next random divisions of features are taken into consideration when nodes split. During training, every tree in a random forest prototype understands from an arbitrary sample of the data points. The conclusive predictions of random forest are done by taking the average of the predictions of each distinctive tree. In order for random forest to make decisions like humans, it needs to learn things and for this learning, it needs to have a humongous amount of information in its training set. After learning from the training set, it is able to process the information and they can classify the given set of data into a predefined class.

\subsection{Model Representation} 
The brick of a random forest is a decision tree which is an intuitive model. For example, a sequence of yes-no queries asked about the data gradually leading to a predicted class is a decision tree. It is an interpretable model because it constructs categorizations almost as humans do. This tree is constructed by shaping the questions via splitting of nodes as in Fig. 2. The answers from nodes lead to a high order reduction in Gini Impurity. Decision trees form nodes comprising a higher proportion of data points from a unit class by taking out values in the features that vividly divide the data into different classes.

\begin{figure}[h]
   \centering
    \includegraphics[width=0.75\textwidth]{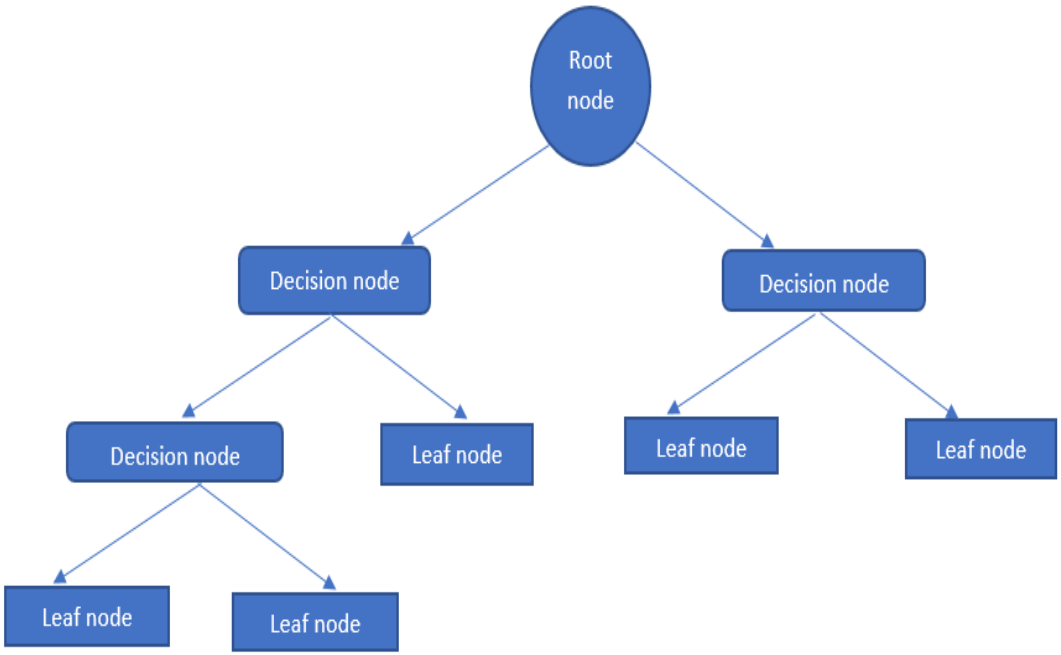}
    \caption{Predictive model’s network representation }
    \label{fig:my_label6}
\end{figure}

\subsection{Gini Impurity} 
The weighted average of Gini Impurity keeps on decreasing as we proceed further to the lower levels of the tree. The Gini Impurity of a node is the prospect that a sample preferred randomly in a node would be wrongly categorized if it got categorized by the allocation of experiments in the node. Eq. (3) defines the formula for finding a Gini Impurity of any node $n$. At every node, the tree looks out for all the characteristics for finding the value to split on that can result in the severe drop in Gini Impurity. Then it reiterates the split process in a greedy, recursive manner until it reaches a maximum depth or each node has only samples from one class.
 
\begin{equation}
    I_G(n)=1-\Sigma^J_{i=1}(p_i)^2
\end{equation}

\section{Performance Evaluation}
The implementation of the proposed model is described in this section followed by accuracy results.
\subsection{Experimental Set-up and Benchmark Dataset}
For implementing the model, the Pyspark platform is used with python language. Pyspark is used for structured and semi-structured data. It can also read data from various data sources having different file formats with the help of an API.  The dataset is attained from the deep learning repository which is from the University of California(UCI)\cite{UCI}. The patient’s ages were ranging from 33 to 87 who was suffering from Parkinson’s disease. The task is of classification and the dataset is of multivariate characteristics. The number of attributes is 754 and it has no missing values. The class values are 0 and 1, depicting the occurrence of disease or not. The database contains information like age etc. Forecast Accuracy, sensitivity, specificity, precision, and F1 Score are evaluated for the random forest classifier to validate the performance. The evaluation of these metrics for the proposed scheme is done using a confusion matrix.  

\subsection{Experimental Results}
To evaluate the model, a 2×2 confusion matrix is used. The confusion matrix of size 2 × 2  is the measure of correct and wrong predictions which are abridged with count values. The model is evaluated using the following various performance metrics that are computed from the confusion matrix where $TN$ reflects a true negative case, where $TP$ reflects truly-positive cases, $FP$ shows false-positive cases and $FN$ holds the false-negative cases.

\begin{itemize}
\item \textbf{Sensitivity:} Sensitivity can be defined as the model’s ability to accurately detect the patients who actually have the disease. 
\begin{equation}
    Sensitivity=\frac{TP}{TP+FN}
\end{equation}
\item \textbf{Specificity:} Specificity can be defined as the model’s ability to accurately identify the people who are healthy and who don’t have the condition. 
\begin{equation}
    Specificity=\frac{TN}{TN+FP}
\end{equation}
\item \textbf{Accuracy:} Accuracy is delineated as to the frequency of correct result predictions out of the total predictions from the dataset. 
\begin{equation}
    Accuracy=\frac{\text{\#Number of correct predictions}}{\text{\#Total number of predictions}}
\end{equation}
\item \textbf{Precision:} Precision tells us how precise the prototype is out of the predicted positive and what number of those are really positive.
\begin{equation}
    Precision=\frac{TP}{TP+FP}
\end{equation}
\item \textbf{F1 Score:}  This score is the harmonic mean between two vital stats which are sensitivity and precision.
\begin{equation}
    F1 Score=\frac{2TP}{2TP+FP+FN}
\end{equation}
\item \textbf{ROC Curve:}  Receiver’s Operating characteristics curve is a performance metric for evaluating a classification model’s performance. It is a probability curve that is plotted as TPR on the vertical dimension Y-axis and FPR on the horizontal dimension X-axis. 
\end{itemize} 

The performance in terms of accuracy, sensitivity, specificity, precision, and F1 Score of PCA-RF model without PCA and with PCA is depicted in Table 1. It can be seen that the PCA-RF model secures 89.9\% and 76.7\% accuracy, 70.2\% and 55.6\% sensitivity, 96.5\% and 80.6\% specificity, 70.2\% and 35.1\% precision, 77.7\%, and 43\% F1 Score without and with PCA respectively. It is observed that the performance without PCA is better compared to with PCA for every parameter accuracy, sensitivity, specificity, precision, and F1 Score since the performance greatly relies on the amalgamation of the feature reduction technique with the proposed classifier model. 

\begin{table}[h]
\caption{Performance metrıcs for PCA-RF model}
\centering
 \begin{tabular}{|c | c | c |} 
 \hline
 \textbf{Performance Metrics} & \textbf{Without PCA} & \textbf{With PCA} \\ 
 \hline
 Accuracy  & 89.867   & 76.651  \\ 
 \hline
 Sensitivity & 70.175 & 55.555 \\
 \hline
 Specificity & 96.470 & 80.628 \\
 \hline
 Precision & 70.175 & 35.087 \\
 \hline
 F1 Score & 77.669 & 43.010 \\
 \hline
 \end{tabular}
\label{Table 1.}
\end{table}

we have plotted the ROC curve for Parkinson’s disease that is depicted in  Fig.3. The figure shows the deflection of the model curve from the baseline where the horizontal axis (x-axis) demonstrates a false positive rate and the vertical axis (y-axis) depicts the true positive rate. It can be seen that the true positive rate of the proposed model is high which validates its performance.

\begin{figure}[H]
	\centering
	\includegraphics[width=0.75\textwidth]{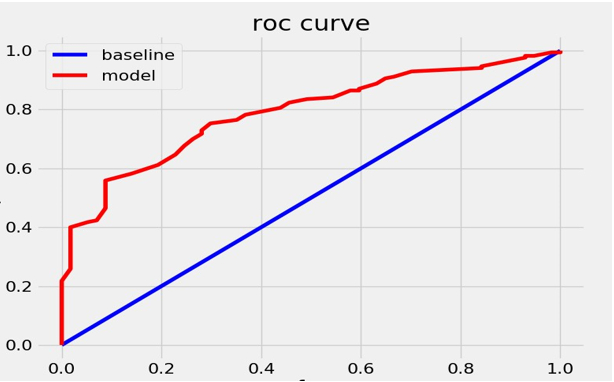}
	\caption{ROC curve of the proposed model}
	\label{fig:my_label}
\end{figure}

\subsection{Comparative Analysis}
Random forest classifier, when compared with Artificial Neural Network, shows a great deflection in the accuracy. Random forest works well with tabular data and being an intuitive model, it works on decision trees. At every step, each attribute plays a significant role in arriving at the desired result. Whereas Artificial Neural Network is an advanced algorithm of machine learning where a group of attributes is selected that participate in the classification process to predict the result. ANN has a large number of layers, each layer’s output being input to the next layer. The accuracy of both the models is demonstrated in Table 2 when applied with and without a feature reduction technique. Table 2 shows the performance in terms of accuracy, sensitivity, specificity, precision, F1 score, and the comparison among ANN and Random Forest with PCA and without PCA respectively.
The depicted results show a diverge deflection when applied with and without PCA. When applied with PCA the random forest classifier has low performance compared to ANN. Whereas, in the case of Artificial Neural Network when PCA has applied the accuracy greatly increases.
While random forest classifier has high performance whereas ANN as comparatively low performance in case of without PCA. It depends on the importance of features and the dimension of the dataset. Dimensionality reduction is required for removing the redundancy in data that reduces time and storage. It becomes easy to visualize the data by reducing it to a low dimension. 

\begin{table}[h]
\caption{Comparison among ANN and Random Forest}
\centering
 \begin{tabular}{|c | c | c |c | c |} 
 \hline
 \multirow{2}{*}{\textbf{Metrics}} & \multicolumn{2}{c}{with PCA} & \multicolumn{2}{|c}{without PCA} \\\cline{2-5}
  & \textbf{ANN} & \textbf{Random Forest} & \textbf{ANN} & \textbf{Random Forest} \\ 
 \hline
 Accuracy  & 97.354 & 76.651  & 79.470 & 89.534  \\ 
 \hline
 Sensitivity & 95.454 & 55.555 & 64.516 & 70.175 \\
 \hline
 Specificity & 97.931 & 80.628 & 83.333 & 96.470 \\
 \hline
 Precision & 93.333 & 35.087 & 50.000 & 70.175 \\
 \hline
 F1 Score & 94.382 & 43.010 & 56.338 & 77.669 \\
 \hline
 
 \end{tabular}
 
\label{Table 1.}
\end{table}

\section{Conclusion and Future Scope}
The work elaborated in this paper is based on the evaluation of Random forest classification on the particular dataset which is a high dimensional data containing 754 attributes. In terms of tabular data, random forest works well and so far the particular data we have opted for this classifier is the most appropriate selection. The paper also deals with how accuracy deflects when we pass the dataset to the proposed random forest model and when passed to the artificial neural network along with PCA. There is a visible difference between both models. Accuracy is high for the ANN model since a feature reduction technique is used along with it. It depends on the classifier model used along with the feature reduction technique that ensures good accuracy. Therefore, the accuracy of the proposed model is approximately 90\%.
The paper deals with the prediction of only one type of disease, so further diseases can be predicted simultaneously for early detection. Furthermore, the accuracy of the various models can be contrasted and the best classifier model can be selected for that particular disease.

\input{main.bbl}

\end{document}